\documentclass{article}
\usepackage{ijcai16}
\usepackage{boxedminipage}
\usepackage{times}
\usepackage{url}
\usepackage{amsmath}
\usepackage{amsfonts}
\usepackage{amssymb}
\usepackage[noend]{algorithm2e}
\usepackage{graphicx} % Allows including images
\usepackage{svg}
\usepackage{xifthen}

\title{Real-Time Web Scale Event Summarization Using Sequential Decision 
    Making}
\author{Chris Kedzie \\ 
Columbia University  \\
Dept. of Computer Science\\ 
kedzie@cs.columbia.edu 
\And
Fernando Diaz\\
    Microsoft Research\\
    fdiaz@microsoft.com
    \And 
    Kathleen McKeown\\
Columbia University  \\
Dept. of Computer Science\\ 
kathy@cs.columbia.edu 
}

\newcommand{\doc}{X}

\newcommand{\loss}{\ell}

\newcommand{\spx}{\mathbf{x}}
\newcommand{\spxi}{x}
\newcommand{\spy}{\mathbf{a}}
\newcommand{\spyi}{a}

\newcommand{\Feat}{\Phi}
\newcommand{\Actions}{\mathcal{A}}

\newcommand{\action}{a}
\newcommand{\States}{\mathcal{S}}
\newcommand{\states}{\States}
\newcommand{\state}{s}
\newcommand{\Train}{\Gamma}

\newcommand{\query}{q}
\newcommand{\queries}{\mathcal{Q}}
\newcommand{\oracle}{\pi^*}
\newcommand{\model}{\tilde{\pi}}
\newcommand{\rolloutpolicy}{\pi^{\text{o}}}

\newcommand{\modeli}{\model_i}
\newcommand{\modeliplusone}{\model_{i+1}}
\newcommand{\modelzero}{\model_0}

\newcommand{\stream}{\spx}

\newcommand{\streamt}{\stream_{:t}}
\newcommand{\decisions}{\spy}

\newcommand{\decisionst}{\decisions_{:t-1}}

\begin{document}

\maketitle

\begin{abstract}

 We present a system based on sequential decision making for the online 
 summarization of massive document streams, such as those found on the web.  
 Given an event of interest (e.g. ``Boston marathon bombing''), our system is 
 able to filter the stream for relevance and produce a series of short text 
 updates describing the event as it unfolds over time. Unlike previous work, 
 our approach is able to jointly model the relevance, comprehensiveness, 
 novelty, and timeliness required by time-sensitive queries.  We demonstrate 
 a 28.3\% improvement in summary $F_1$ and a 43.8\% improvement in
 time-sensitive 
 $F_1$ metrics. 

\end{abstract}

\section{Introduction}
\label{sec:introduction}
  Tracking unfolding news at web-scale continues to be a challenging task.
 Crisis informatics, monitoring of breaking news, and intelligence tracking
 all have difficulty in identifying new, relevant information within the
 massive quantities of text that appear online each second.  One broad need
 that has emerged is the ability to provide realtime event-specific updates of
 streaming text data which are timely, relevant, and comprehensive while
 avoiding redundancy.  
 
  Unfortunately, many approaches have adapted standard automatic
 multi-document summarization techniques that are  inadequate for web scale
 applications.  Typically, such systems assume full retrospective access to
 the documents to be summarized, or that at most a handful of updates to the
 summary will be made \cite{dang2008overview}.  Furthermore, evaluation of
 these systems has assumed reliable and relevant input, something missing in
 an inconsistent, dynamic, noisy stream of web or social media data.  As a
 result, these systems are poor fits for most real world applications. 
 
  In this paper, we present a novel streaming-document summarization system
 based on sequential decision making.  Specifically, we adopt the ``learning
 to search'' approach, a technique which adapts methods from reinforcement
 learning for structured prediction problems
 \cite{daume2009search,ross2010reduction}.  In this framework, we cast
 streaming summarization as a form of greedy search and train our system to
 imitate the behavior of an oracle summarization system.  
 
  Given a stream of sentence-segmented news webpages and an event query  (e.g.
 ``Boston marathon bombing''), our system monitors the stream to detect
 relevant, comprehensive, novel, and timely content.  In response, our
 summarizer produces a series of short text updates describing  the event as
 it unfolds over time.  We present an example of our realtime update stream in
 Figure~\ref{fig:system-summary}.  We evaluate our system in a crisis
 informatics setting on a diverse set of event queries, covering severe
 storms, social unrest, terrorism, and large accidents. We demonstrate a
 28.3\% improvement in summary $F_1$ and a 43.8\% improvement in
 time-sensitive $F_1$ metrics against several state-of-the-art baselines. 
 
\begin{figure}
  \begin{boxedminipage}{\columnwidth}
    \center
    \footnotesize
    \begin{tabular}{p{.95\columnwidth}} % p{.15\columnwidth}p{.77\columnwidth}}
    \textit{4:19 p.m.} --  Two explosions shattered the euphoria of the Boston
 Marathon finish line on Monday, sending authorities out on the course to 
 carry off the injured while the stragglers were rerouted away...\\[0.5em] 
 \textit{4:31 p.m.} --  Police in New York City and London are stepping up 
 security following explosions at the Boston Marathon. \\[0.5em] 
 \textit{4:31 p.m.} -- A senior U.S. intelligence official says two more 
 explosive devices have been found near the scene of the Boston marathon where
 two bombs detonated earlier. \\[0.5em] 
 \textit{5:10 p.m.} -- Several candidates for Massachusetts’ Senate 
 special election have suspended campaign activity in response to the 
 explosions... \\[0.5em] 
    \end{tabular}
  \end{boxedminipage}
  \caption{Excerpt of summary for the query `Boston marathon bombing'
           generated from an input stream.}
  \label{fig:system-summary}
\end{figure}

\section{Related Work}
\label{sec:relatedwork}
  Multi-document summarization (MDS) has long been studied by the natural
 language processing community. We focus specifically on extractive
 summarization, where the task is to take a collection of text and select some
 subset of sentences from it that adequately describes the content subject to
 some budget constraint (e.g. the summary must not exceed $k$ words). For a
 more in depth survey of the field, see \cite{nenkova2012survey}.
 
  Because labeled training data is often scarce, unsupervised approaches to
 clustering and ranking predominate the field. 
 
  Popular approaches involve ranking sentences by various notions of input
 similarity or graph centrality \cite{radev2000centroid,erkan2004lexrank}. 
 
  Other ranking based methods use coverage of topic signatures
 \cite{lin2000automated}, or KL divergence between input/summary word
 distributions \cite{haghighi2009exploring} as the ranking, possibly adding
 some diversity penalty to ensure broader coverage. 
 
  Update summarization research has primarily focused on producing a 100 word
 summary from a set of 10 documents, assuming the reader is familiar with a
 different initial set of 10 documents \cite{dang2008overview}.  Generally the
 approaches to update summarization have adapted the above techniques. Top
 performers at the 2011 Text Analysis Conference (the last year update
 summarization was a task) made use of graph ranking algorithms, and topic
 model/topic-signature style importance estimates
 \cite{du2011decayed,mason2011extractive,conroy2011classy}. 
 
  Streaming or temporal summarization was first explored in the context of
 topic detection and tracking \cite{allan2001temporal} and more recently at
 the Text Retrieval Conference (TREC) \cite{aslam2013trec}. Top performers at
 TREC included an affinity propagation clustering approach
 \cite{kedziepredict} and a ranking/MDS system combination method
 \cite{mccreadie2014incremental}.  Both methods are unfortunately constrained
 to work in hourly batches, introducing potential latency. Perhaps most
 similar to our work is that of \cite{guo2013updating} which iteratively fits
 a pair of regression models to predict ngram recall and precision of
 candidate updates to a model summary. However, their learning objective fails
 to account for errors made in subsequent prediction steps. 

\section{Problem Definition}
\label{sec:problem}
  A streaming summarization task is composed of a brief text query $\query$,
 including a categorical event type (e.g. `earthquake', `hurricane'), as well
 as a document stream $[\doc_0,\doc_1,\ldots]$.  In practice, we assume that
 each document is segmented into a sequence of sentences and we therefore
 consider a sentence stream $[\spxi_0,\spxi_1,\ldots]$.  A streaming
 summarization algorithm then selects or skips each sentence $\spxi$ as it is
 observed such that the end user is provided a filtered stream of sentences
 that are relevant, comprehensive, low redundancy, and timely (see
 Section~\ref{sec:metrics}).  We refer to the selected sentences as
 \textit{updates} and collectively they make up an  \textit{update summary}.
 We show a fragment of an update summary for the query `Boston marathon
 bombing' in Figure~\ref{fig:system-summary}.  

\section{Streaming Summarization as Sequential Decision Making}

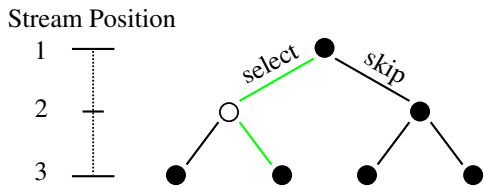
\begin{figure}
  \center
%% Creator: Inkscape inkscape 0.48.4, www.inkscape.org
%% PDF/EPS/PS + LaTeX output extension by Johan Engelen, 2010
%% Accompanies image file 'search.pdf' (pdf, eps, ps)
%%
%% To include the image in your LaTeX document, write
%%   \input{<filename>.pdf_tex}
%%  instead of
%%   \includegraphics{<filename>.pdf}
%% To scale the image, write
%%   \def\svgwidth{<desired width>}
%%   \input{<filename>.pdf_tex}
%%  instead of
%%   \includegraphics[width=<desired width>]{<filename>.pdf}
%%
%% Images with a different path to the parent latex file can
%% be accessed with the `import' package (which may need to be
%% installed) using
%%   \usepackage{import}
%% in the preamble, and then including the image with
%%   \import{<path to file>}{<filename>.pdf_tex}
%% Alternatively, one can specify
%%   \graphicspath{{<path to file>/}}
%% 
%% For more information, please see info/svg-inkscape on CTAN:
%%   http://tug.ctan.org/tex-archive/info/svg-inkscape
%%
\begingroup%
  \makeatletter%
  \providecommand\color[2][]{%
    \errmessage{(Inkscape) Color is used for the text in Inkscape, but the package 'color.sty' is not loaded}%
    \renewcommand\color[2][]{}%
  }%
  \providecommand\transparent[1]{%
    \errmessage{(Inkscape) Transparency is used (non-zero) for the text in Inkscape, but the package 'transparent.sty' is not loaded}%
    \renewcommand\transparent[1]{}%
  }%
  \providecommand\rotatebox[2]{#2}%
  \ifx\svgwidth\undefined%
    \setlength{\unitlength}{178.49460449bp}%
    \ifx\svgscale\undefined%
      \relax%
    \else%
      \setlength{\unitlength}{\unitlength * \real{\svgscale}}%
    \fi%
  \else%
    \setlength{\unitlength}{\svgwidth}%
  \fi%
  \global\let\svgwidth\undefined%
  \global\let\svgscale\undefined%
  \makeatother%
  \begin{picture}(1,0.35907832)%
    \put(0,0){\includegraphics[width=\unitlength]{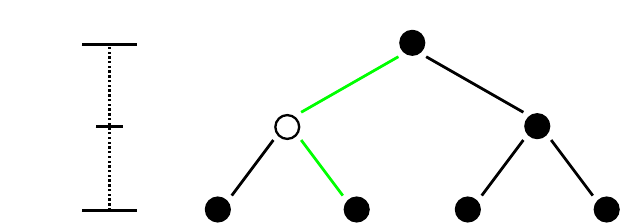}}%
    \put(-0.00356303,0.37186889){\color[rgb]{0,0,0}\makebox(0,0)[lt]{\begin{minipage}{1.32653733\unitlength}\raggedright Stream Position \end{minipage}}}%
    \put(0.05207008,0.30530006){\color[rgb]{0,0,0}\makebox(0,0)[lt]{\begin{minipage}{0.06721748\unitlength}\raggedright 1\end{minipage}}}%
    \put(0.05326497,0.1819457){\color[rgb]{0,0,0}\makebox(0,0)[lt]{\begin{minipage}{0.06721748\unitlength}\raggedright 2\end{minipage}}}%
    \put(0.05264783,0.04993889){\color[rgb]{0,0,0}\makebox(0,0)[lt]{\begin{minipage}{0.06721748\unitlength}\raggedright 3\end{minipage}}}%
    \put(0.483084,0.24693812){\color[rgb]{0,0,0}\rotatebox{27.31235445}{\makebox(0,0)[lt]{\begin{minipage}{0.18248248\unitlength}\raggedright select\end{minipage}}}}%
    \put(0.76504897,0.29214953){\color[rgb]{0,0,0}\rotatebox{-29.99982322}{\makebox(0,0)[lt]{\begin{minipage}{0.13269863\unitlength}\raggedright skip\end{minipage}}}}%
  \end{picture}%
\endgroup%
  
  \caption{Search space for a stream of size two. The depth of the tree 
           corresponds to the position in the stream. Left branches indicate 
           selecting the current sentence as an update. Right branches skip 
           the current sentence. The path in green corresponds to one 
           trajectory through this space consisting of a select sentence one, 
           then skip sentence 2. The state represented by the hollow dot 
           corresponds to the stream at sentence position 2 with the update 
           summary containing sentence 1.}
  \label{fig:search}
\end{figure}

  We could na{\"i}vely treat this problem as classification and predict which
 sentences to select or skip. However, this would make it difficult to take
 advantage of many features (e.g. sentence novelty w.r.t. previous updates).
 What is more concerning, however, is that the classification objective for
 this task is somewhat ill-defined: successfully predicting select on one
 sentence changes the true label (from select to skip) for sentences that
 contain the same information but occur later in the stream.
 
  In this work, we pose streaming summarization as a greedy search over a
 binary branching tree where each level corresponds to a position in the
 stream (see Figure \ref{fig:search}).  The height of the tree corresponds to
 the length of stream.  A path through the tree is determined by the system
 select and skip decisions.
 
  When treated as a sequential decision making problem, our task reduces to
 defining a policy for selecting a sentence based on its properties as well as
 properties of its ancestors (i.e. all of the observed sentences and previous
 decisions).  The union of properties--also known as the features--represents
 the current state in the decision making process. 
 
  The feature representation provides state abstraction both within a given
 query's search tree as well as to states in other queries' search trees, and
 also allows for complex interactions between the current update summary,
 candidate sentences, and stream dynamics unlike the classification approach.
  
  In order to learn an effective policy for a query $\query$, we can take one
 of several approaches.  We could use a simulator to provide feedback to a
 reinforcement learning algorithm.  Alternatively, if provided access to an
 evaluation algorithm at training time, we can simulate (approximately)
 optimal decisions. That is, using the training data, we can define an
 \emph{oracle policy} that is able to omnisciently determine which sentences
 to select and which to skip.  Moreover, it can make these determinations by
 starting at the root or at an arbitrary node in the tree, allowing us to
 observe optimal performance in states unlikely to be reached by the oracle.
 We adopt \textit{locally optimal learning to search} to learn our model from
 the oracle policy \cite{chang2015learning}.  
 
  In this section, we will begin by describing the learning algorithm
 abstractly and then in detail for our task.  We will conclude with details on
 how to train the model with an oracle policy.  

\begin{algorithm}[]
  \RestyleAlgo{boxruled}
  \LinesNumbered
  \KwIn{$\{\spx_{\query}, \oracle_{\query} \}_{\query\in\queries}$, number of 
        iterations $N$, and a mixture parameter $\beta \in (0,1)$ for 
        roll-out.}
  \KwOut{$\model$}
  Initialize $\modelzero$ \\
  $i\gets 0$\\
  \For{$n \in \{1,2,\ldots, N\}$}{
    \For{$\query \in \queries$}{
      $\Train \gets \emptyset$\\    
      \For{$t \in \{0,1,\ldots,T-1 \}$}{
        Roll-in by executing $\modeli$ for $t$ rounds and reach $s_t$. \\
        \For{$a \in \Actions(s_t)$}{
            Let $\rolloutpolicy = \oracle_{\query}$ with probability 
            $\beta$, else $\modeli$. \\
            Compute $c_t(a)$ by rolling out with $\rolloutpolicy$\\
            $\Train \gets \Train \cup \{\langle \Feat(\state_t),\action,
            c_t(\action)\rangle\}$}}
      $\modeliplusone \gets \textrm{Update Cost-Sensitive Classifier}(
          \modeli, \Train)$\\
      $i\gets i+1$}}
  Return $\model_i$.
  \caption{Locally optimal learning to search.} 
  \label{fig:lols}
\end{algorithm}

\subsection{Algorithm}
\label{sec:algorithm}

 In the induced search problem, each search state $s_t$ corresponds to
 observing the first $t$ sentences in the stream $x_1,\ldots, x_t$ and a
 sequence of $t-1$ actions $\action_1, \ldots, \action_{t-1}$.  For all
 states $\state \in \states$, the set of actions is $\action \in \{0,1\}$
 with $1$ indicating we add the $t$-th sentence to our update summary, and
 $0$ indicating we ignore it. For simplicity, we assume a fixed length stream
 of size $T$ but this is not strictly necessary. From each input stream,
 $\spx = \spxi_1, \ldots, \spxi_T$, we produce a corresponding output $\spy
 \in \{0,1\}^T$.  We use $\streamt$ to indicate the first $t$ elements of
 $\spx$.  

 For a training query $\query$, a reference policy $\oracle_q$ can be
 constructed from the training data.  Specifically, with access to the
 relevance and novelty of every $x_i$ in the stream, we can omnisciently make
 a decision whether to select or skip based on a long term metric (see
 Section~\ref{sec:metrics}). The goal then is to learn a policy $\model$ that
 imitates the reference well across a set of training queries $\queries$. We
 encode each state as a vector in $\mathbb{R}^d$ with a feature function
 $\Feat$ and our learned policy is a mapping $\model : \mathbb{R}^d \to
 \Actions$ of states to actions.  
 
 We train $\model$ using locally optimal learning to
 search~\cite{chang2015learning}, presented in Algorithm~\ref{fig:lols}.  The
 algorithm operates by iteratively updating a cost-sensitive classifier.  For
 each training query, we construct a query-specific training set $\Train$ by
 simulating the processing of the training input stream $\spx_q$.  The
 instances in $\Train$ are triples comprised of a feature vector derived from
 the current state $\state$, a candidate action $\action$, and the cost
 $c(a)$ associated with taking action $\action$ in state $\state$.
 Constructing $\Train$ consists of (1) selecting states and actions, and (2)
 computing the cost for each state-action pair.
 
 The number of states is exponential in $T$, so constructing $\Train$ using
 the full set of states may be computationally prohibitive.  Beyond this, the
 states in $\Train$ would not be representative of those visited at test
 time.  In order to address this, we sample from $S$ by executing the current
 policy $\model$ throughout the training simulation, resulting in $T$ state
 samples for $\Train$, (lines 5-12).
 
 Given a sampled state $\state$, we need to compute the cost of taking
 actions $\action \in \{0, 1\}$.  With access to a query-specific oracle,
 $\oracle_q$, we can observe it's preferred decision at $\state$ and penalize
 choosing the other action.  The magnitude of this penalty is proportional to
 the difference in expected performance between the oracle decision and the
 alternative decision.   The performance of a decision is derived from a loss
 function $\loss$, to be introduced in Section~\ref{sec:oracle}.
 Importantly, our loss function is defined over a complete update summary,
 incorporating the implications of selecting an action on future decisions.
 Therefore, our cost needs to incorporate a sequence of decisions after
 taking some action in state $\state$.  The algorithm accomplishes this by
 \emph{rolling out} a policy after $\action$ until the stream has been
 exhausted (line 10).  As a result, we have a prefix defined by $\model$, an
 action, and then a suffix defined by the roll out policy.  In our work,
 we use a mixture policy that combines both the current model $\model$ and
 the oracle $\oracle$ (line 9). This mixture policy encourages learning from
 states that are likely to be visited by the current learned policy but not
 by the oracle.
 
 After our algorithm has gathered $\Train$ for a specific $\query$ using
 $\model_i$, we train on the data to produce $\model_{i+1}$.   Here
 $\model_i$ is implemented as a cost-sensitive classifier, i.e. a linear
 regression of the costs on features and actions; the natural policy is to
 select the action with lowest predicted cost.  With each query, we update
 the regression with stochastic gradient descent on the newly sampled (state,
 action, cost) tuples (line 12).  We repeat this process for $N$ passes over
 all queries in the training set.

 In the following sections, we specify the feature function $\Feat$, the loss
 $\loss$, and our reference policy  $\pi^*$.

\subsection{Features}

 As mentioned in the previous section, we represent each state as a feature
 vector.  In general, at time $t$, these features are functions of the
 current sentence (i.e. $\spxi_t$), the  stream history  (i.e. $\streamt$),
 and/or the  decision history ($\decisionst$).  We refer to features only
 determined by $\spxi_t$ as static features and all others as dynamic
 features.  \footnote{\scriptsize We have attempted to use a comprehensive
     set of static features used in previous summarization systems.  We omit
 details for space but source code is available at:
 \texttt{https://github.com/kedz/ijcai2016}}

\subsubsection{Static Features}

 \textbf{Basic Features } Our most basic features look at the length in words
 of a sentence, its position in the document, and the ratio of specific named
 entity tags to non-named entity tokens.  We also compute the average number
 of sentence tokens that match the event query words and synonyms using
 WordNet.

 \textbf{Language Model Features } Similar to \cite{kedziepredict}, we
 compute the average token log probability of the sentence on two language
 models: i) an event type specific language model and ii) a general newswire
 language model.  The first language model is built from Wikipedia articles
 relevant to the event-type domain. The second model is built from the New
 York Times and Associate Press sections of the Gigaword-5 corpus
 \cite{graff2003english}.

 \textbf{Single Document Summarization Features } These features are computed
 using the current sentence's document as a context and are also commonly
 used as ranking features in other document summarization systems. Where a
 similarity or distance is need, we use either a tf-idf bag-of-words or
 $k$-dimensional latent vector representation. The latter is derived by
 projecting the former onto a $k$-dimensional space using the weighted
 textual matrix factorization method \cite{guo:wtmf}.

 We compute \textsc{SumBasic} features \cite{nenkova:sumbasic}: the average
 and sum of unigram probabilities in a sentence.  We compute the arithmetic
 and geometric means of the sentence's cosine distance to the other sentences
 of the document \cite{guo2013updating}. We refer to this quantity as novelty
 and compute it with both vector representations. We also compute the
 centroid rank \cite{radev2000centroid} and LexRank of each sentence
 \cite{erkan2004lexrank}, again using both vector representations.
 
 \textbf{Summary Content Probability} For a subset of the stream sentences we
 have manual judgements as to whether they match to model summary content or
 not (see Sec. \ref{sec:data}, Expanding Relevance Judgments). We use this
 data (restricted to sentences from the training query streams), to train a
 decision tree classifier, using the sentences' term ngrams as classifier
 features. As this data is aggregated across the training queries, the
 purpose of this classifier is to capture the importance of general ngrams
 predictive of summary worthy content.  

 Using this classifier, we obtain the probability that the current sentence
 $\spxi_t$ contains summary content and use this as a model feature.

\subsubsection{Dynamic Features}

 \textbf{Stream Language Models} We maintain several unigram language models
 that are updated with each new document in the stream. Using these counts,
 we compute the sum, average, and maximum token probability of the non-stop
 words in the sentence. We compute similar quantities restricted to the
 \textit{person}, \textit{location}, and \textit{organization} named
 entities.

 \textbf{Update Similarity} The average and maximum cosine similarity of the
 current sentence to all previous updates is computed under both the tf-idf
 bag-of-words and latent vector representation. We also include indicator
 features for when the set of updates is empty (i.e. at the beginning of a
 run) and when either similarity is 0.

 \textbf{Document Frequency} We also compute the hour-to-hour percent change
 in document frequency of the stream. This feature helps gauge breaking
 developments in an unfolding event. As this feature is also heavily affected
 by the daily news cycle (larger average document frequencies in the morning
 and evening) we compute the 0-mean/unit-variance of this feature using the
 training streams to find the mean and variance for each hour of the day.

 \textbf{Feature Interactions} Many of our features are helpful for
 determining the importance of a sentence with respect to its document.
 However, they are more ambiguous for determining importance to the event as
 a whole. For example, it is not clear how to compare the document level
 PageRank of sentences from different documents. To compensate for this, we
 leverage two features which we believe to be good global indicators of
 update selection: the summary content probability and the document
 frequency.  These two features are proxies for detecting (1) a good summary
 sentences (regardless of novelty with respect to other previous decisions)
 and (2) when an event is likely to be producing novel content. We compute
 the conjunctions of all previously mentioned features with the summary
 content probability and document frequency separately and together.

\subsection{Oracle Policy and Loss Function}
\label{sec:oracle}
 
 Much of the multi-document summarization literature employs greedy selection
 methods.  We adopt a greedy oracle that selects a sentence if it improves
 our evaluation metric (see Section~\ref{sec:metrics}).  

 We design our loss function to penalize policies that severely over- or
 under-generate. Given two sets of decisions, usually one from the oracle and
 another from the candidate model, we define the loss as the complement of
 the Dice coefficient between the decisions,
\begin{align*}
  \loss(\decisions, \decisions') = 1 - 2 \times 
    \frac{ \sum_i \spyi_i \spyi^\prime_i }{ \sum_i \spyi_i + \spyi^\prime_i }.
\end{align*}

 This encourages not only local agreement between policies (the numerator of
 the second term) but that the learned and oracle policy should generate
 roughly the same number of updates (the denominator in the second term).

\section{Materials and Methods}
\subsection{Data}
\label{sec:data}
 We evaluate our method on the publicly available TREC Temporal Summarization
 Track data.\footnote{\scriptsize\url{http://www.trec-ts.org/}}  This data is
 comprised of three parts.  

 The \emph{corpus} consists of a 16.1 terabyte set of 1.2 billion timestamped
 documents crawled from the web between October, 2011 and February 2013
 \cite{frank2012building}.  The crawl includes news articles, forum data,
 weblogs, as well as a variety of other crawled web
 pages.\footnote{\scriptsize\url{http://streamcorpus.org/}}  

 The  \emph{queries} consist of a set of 44 events which occurred during the
 timespan of the corpus.  Each query has an associated time range to limit
 the experiment to a timespan of interest, usually around two weeks.  In
 addition, each query is associated with an `event category' (e.g.
 `earthquake', `hurricane'). Each query is also associated with an ideal
 summary, a set of short, timestamped textual descriptions of facts about the
 event.  The items in this set, also known as \emph{nuggets}, are considered
 the completed and irreducible sub-events associated with the query. For
 example, the phrases ``multiple people have been injured'' and ``at least
 three people have been killed'' are two of the nuggets extracted for the
 query `Boston marathon bombing'. On average, 73.35 nuggets were extracted
 for each event.

 The \emph{relevance judgments} consist of a sample of sentences pooled from
 participant systems, each of which has been manually assessed as related to
 one or more of a query's nuggets or not.  For example,  the following
 sentence, ``Two explosions near the finish line of the Boston Marathon on
 Monday killed three people and wounded scores,'' matches the nuggets
 mentioned above.  The relevance judgments can be used to compute evaluation
 metrics (Section~\ref{sec:metrics}) and, as a result, to also define our
 oracle policy (Section~\ref{sec:oracle}).

 \subsubsection{Expanding Relevance Judgments}

 Because of the large size of the corpus and the limited size of the sample,
 many good candidate sentences were not manually reviewed.  After aggressive
 document filtering (see below), less than 1\% of the sentences received
 manual review.  In order to increase the amount of data for training and
 evaluation of our system, we augmented the manual judgements with automatic
 or ``soft'' matches. A separate gradient boosting classifier was trained for
 each nugget with more than 10 manual sentence matches.  Manually matched
 sentences were used as positive training data and an equal number of
 manually judged non-matching sentences were used as negative examples.
 Sentence ngrams (1-5), percentage of nugget terms covered by the sentence,
 semantic similarity of the sentence to nugget were used as features, along
 with an interaction term between the semantic similarity and coverage. When
 augmenting the relevance judgments with these nugget match soft labels, we
 only include those that have a probability greater than 90\% under the
 classifier. Overall these additional labels increase the number of matched
 sentences by 1600\%.
 
 For evaluation, the summarization system only has access to the query and
 the document stream, without knowledge of any nugget matches (manual or
 automatic).

\subsubsection{Document Filtering} \label{sec:docfilter}

 For any given event query, most of the documents in the corpus are
 irrelevant. Because our queries all consist of news events, we restrict
 ourselves to the news section of the corpus, consisting of 7,592,062
 documents.

 These documents are raw web pages, mostly from local news outlets running
 stories from syndication services (e.g. Reuters), in a variety of layouts.
 In order to normalize these inputs we filtered the raw stream for relevancy
 and redundancy with the following three stage process.

 We first preprocessed each document's raw html using an article extraction
 library.\footnote{\scriptsize\url{https://github.com/grangier/python-goose}}
 Articles were truncated to the first 20 sentences. We then removed any
 articles that did not contain all of the query keywords in the article text,
 resulting in one document stream for each query.  Finally, documents whose
 cosine similarity to any previous document was $> .8$ were removed from the
 stream.

\begin{figure*}
    \center
\begin{tabular}{ l | l l l | l l l | l }
   &\multicolumn{3}{c|}{unpenalized}&\multicolumn{3}{c|}{latency-penalized}&\\
   & exp. gain     & comp. & $F_1$ & exp. gain     & comp. & $F_1$ & num. updates\\
    \hline
%\textsc{AP}         
%    & $0.083$            & $0.09$                     & $0.078$ 
%    & $0.079$            & $0.095$                    & $0.077$ 
%    & ~~$20.846^{s}$ \\
\textsc{APSal}      
    & $\mathbf{0.119}^c$ & $0.09$                     & $0.094$ 
    & $0.105$            & $0.088$                    & $0.088$ 
    & ~~~~$8.333$ \\
\textsc{Cos}     
    & $0.075$            & $0.176^{s}$              & $0.099$ 
    & $0.095$            & $0.236^{s}$              & $0.128^{s}$ 
    & $145.615^{s,f}$ \\
\textsc{Ls}       
    & $0.097$            & $\mathbf{0.207}^{s,f}$   & $0.112$ 
    & $0.136^{c}$      & $\mathbf{0.306}^{s,c,f}$ & $0.162^{s}$
    & ~~$89.872^{s,f}$ \\
\textsc{LsCos}
    & $0.115^{c,l}$        & $0.189^{s}$ & ${\bf0.127}^{s,c,l}$
    & ${\bf0.162}^{s,c,l}$ & $0.276^{s}$ & ${\bf0.184}^{s,c,l}$
 & ~~$29.231^{s,c}$ \\
\end{tabular}
\caption{
 Average system performance 
 and average number of updates per event.
 Superscripts indicate significant improvements ($p < 0.05$) between the run and
 competing algorithms using the 
  paired randomization test with the Bonferroni correction for multiple 
  comparisons ($s$: \textsc{APSal}, 
 $c$: \textsc{Cos}, $l$: \textsc{Ls}, $f$: \textsc{LsCos}). 
}
\label{fig:results}
\end{figure*}

\subsection{Metrics} \label{sec:metrics} 

 We are interested in measuring a summary's relevance, comprehensiveness,
 redundancy, and latency (the delay in selecting nugget information).  The
 Temporal Summarization Track adopts three principle metrics which we review
 here.  Complete details can be found in the Track's official metrics
 document.\footnote{\scriptsize{
    \url{http://www.trec-ts.org/trec2015-ts-metrics.pdf}}}
 We use the official evaluation code to compute all metrics.

 Given a system's update summary $\decisions$ and our sentence-level
 relevance judgments, we can compute the number of matching nuggets found.
 Importantly, a summary only gets credit for the number of unique matching
 nuggets, \emph{not} the number of matching sentences.  This prevents a
 system from receiving credit for selecting several sentences which match the
 same nugget.  We refer to the number of unique matching nuggets as the
 \emph{gain}.  We can also penalize a system which retrieves a sentence
 matching a nugget far after the timestamp of the nugget.  The
 \emph{latency-penalized gain} discounts each match's contribution to the
 gain proportionally to the delay of the first matching sentence.  

 The gain value can be used to compute latency and redundancy-penalized
 analogs to precision and recall.  Specifically, the \emph{expected gain}
 divides the gain by the number of system updates.  This precision-oriented
 metric can be considered the expected number of new nuggets in a sentence
 selected by the system.  The \emph{comprehensiveness} divides the gain by
 the number of nuggets.  This recall-oriented metric can be considered the
 completeness of a user's information after the  termination of the
 experiment.  Finally, we also compute the harmonic mean of expected gain and
 comprehensiveness (i.e. $F_1$).  We present results using either gain or
 latency-penalized gain in order to better understand system behavior.  

 To evaluate our model, we randomly select five events to use as a
 development set and then perform a leave-one-out style evaluation on the
 remaining 39 events.
   
 \subsection{Model Training}

 Even after filtering,  each training query's document stream is still too
 large to be used directly in our combinatorial search space. In order to
 make training time reasonable yet representative, we downsample each stream
 to a length of 100 sentences. The downsampling is done uniformly over the
 entire stream. This is repeated 10 times for each training event to create a
 total of 380 training streams. In the event that a downsample contains no
 nuggets (either human or automatically labeled) we resample until at least
 one exists in the sample. 

 In order to avoid over-fitting, we select the model iteration for each
 training fold based on its performance (in $F_1$ score of expected gain and
 comprehensiveness) on the development set.

 \subsection{Baselines and Model Variants}

 We refer to our ``learning to search'' model in the results as \textsc{Ls}.
 We compare our proposed model against several baselines and extensions. 

 \textbf{Cosine Similarity Threshold} One of the top performing systems in
 temporal-summarization at TREC 2015 was a heuristic method that only
 examined article first sentences, selecting those that were below a cosine
 similarity threshold to any of the previously selected updates. We
 implemented a variant of that approach using the latent-vector
 representation used throughout this work. The development set was used to
 set the threshold. We refer to this model as \textsc{Cos} (Team
 \textsc{WaterlooClarke} at TREC 2015).
  
 \textbf{Affinity Propagation} The next baseline was a top performer at the
 previous year's TREC evaluations~\cite{kedziepredict}. This system processes
 the stream in non-overlapping windows of time, using affinity propagation
 (AP) clustering \cite{frey2007clustering} to identify update sentences (i.e.
 sentences that are cluster centers). As in the \textsc{Cos} model, a
 similarity threshold is used to filter out updates that are too similar to
 previous updates (i.e. previous clustering outputs).  We use the summary
 content probability feature as the preference or salience parameter.  The time
 window size, similarity threshold, and an offset for the cluster preference
 are tuned on the development set.  We use the authors' publicly available
 implementation and refer to this method as \textsc{APSal}.  

 \textbf{Learning2Search+Cosine Similarity Threshold} In this model, which we
 refer to as \textsc{LsCos}, we run \textsc{Ls} as before, but filter the
 resulting updates using the same cosine similarity threshold method as in
 \textsc{Cos}. The threshold was also tuned on the development set. 

\section{Results} \label{sec:results}

  Results for system runs are shown in Figure \ref{fig:results}.  On average,
 \textsc{Ls} and \textsc{LsCos} achieve higher $F_1$ scores than the baseline
 systems in both latency penalized and unpenalized evaluations. For
 \textsc{LsCos}, the difference in mean $F_1$ score was significant compared
 to all other systems (for both latency settings).
 
  \textsc{APSal} achieved the overall highest expected gain, partially because
 it was the tersest system we evaluated. However, only \textsc{Cos} was
 statistically significantly worse than it on this measure. %Interestingly,
 
  In comprehensiveness, \textsc{Ls} recalls on average a fifth of the nuggets
 for each event. This is even more impressive when  compared to the average
 number of updates produced by each system (Figure \ref{fig:results}); while
 \textsc{Cos} achieves similar comprehensiveness, it takes on average about
 62\% more updates than \textsc{Ls} and almost 400\% more updates than
 \textsc{LsCos}. The output size of \textsc{Cos} stretches the limit of the
 term ``summary,'' which is typically shorter than 145 sentences in length.
 This is especially important if the intended application is negatively
 affected by verbosity (e.g. crisis monitoring).

\begin{figure}
  \center
  \begin{tabular}{ l | l l l |}
    &\multicolumn{3}{c}{latency-penalized}\\
    & exp. gain     & comp. & $F_1$ \\
    \hline
    \small \textsc{Cos}  & $0.095$ & $\mathbf{0.236}$ & $0.128$ \\
    \small \textsc{LsFs}  & $0.164$ & $0.220$ & $0.157$ \\
    \small \textsc{LsCosFs} & 
      $\mathbf{0.207}$ & $0.18~~$ & $\mathbf{0.163}$ \\
  \end{tabular}
  \caption{Average system performance. \textsc{LsFs} and \textsc{LsCosFs} 
           runs are trained and evaluated on first sentences only (like the 
           \textsc{Cos} system). Unpenalized results are omitted for space but
           the rankings are consistent.}
  \label{fig:results-trunc}
\end{figure}

\begin{figure}
  \center
  \begin{tabular}{l|l|l|l|l|l}
    & Miss  &      Miss  &          &    &  \\
    & Lead &      Body &         Empty &   Dupl. & Total \\
    \hline
    \small \textsc{APSal}         
      & 29.6\% & 68.7\% & ~~1.6\% & 0.1\% & 15,986\\
    \small \textsc{Cos}     
      & 17.8\% & 39.4\% & 41.1\% & 1.7\% & 12,873\\
    \small \textsc{LsFs}      
      & 25.4\% & 71.7\% & ~~2.0\% & 0.9\% & 13,088\\
    \small \textsc{LsCosFs}
      & 27.9\% & 70.8\% & ~~1.0\% & 0.2\% & 15,756\\
    \small \textsc{Ls}           
      & 19.6\% & 55.3\% & 19.9\% & 5.1\% & 13,380\\
    \small \textsc{LsCos}    
      & 24.6\% & 66.7\% & ~~7.5\% & 1.2\% & 11,613\\
  \end{tabular}
  \caption{Percent of errors made and total errors on test set.}
  \label{fig:errors}
\end{figure}

%\begin{figure}
%  \center
%  \begin{tabular}{l|l|l|l|l|l}
%    & Miss  &        Miss  &   Empty &  Dupl.   &  \\
%    & Lead \% &      Body \% &    \% &   \% & Total \\
%    \hline
%    \small \textsc{APSal}         
%      & 29.6& 68.7& ~~1.6& 0.1& 15,986\\
%    \small \textsc{Cos}     
%      & 17.8& 39.4& 41.1& 1.7& 12,873\\
%    \small \textsc{LS-FS}      
%      & 25.4& 71.7& ~~2.0& 0.9& 13,088\\
%    \small \textsc{LS-Cos-FS}
%      & 27.9& 70.8& ~~1.0& 0.2& 15,756\\
%    \small \textsc{LS}           
%      & 19.6& 55.3& 19.9& 5.1& 13,380\\
%    \small \textsc{LS-Cos}    
%      & 24.6& 66.7& ~~7.5& 1.2& 11,613\\
%  \end{tabular}
%  \caption{Percent of errors made and total errors on test set.}
%  \label{fig:errors}
%\end{figure}

\section{Discussion} \label{sec:discussion}

  Since \textsc{Cos} only considers the first sentence of each document, it
 may miss relevant sentences below the article's lead. In order to confirm the
 importance of modeling the oracle, we also trained and evaluated the
 \textsc{Ls} based approaches on first sentence only streams. Figure
 \ref{fig:results-trunc} shows the latency penalized results of the first
 sentence only runs.  The \textsc{Ls} approaches still dominate \textsc{Cos}
 and receive larger positive effects from the latency penalty despite also
 being restricted to the first sentence. Clearly having a model (beyond
 similarity) of what to select is helpful. Ultimately we do much better when
 we can look at the whole document.
  
  We also performed an error analysis to further understand how each system
 operates.  Figure \ref{fig:errors} shows the errors made by each system on
 the test streams.  Errors were broken down into four categories. \emph{Miss
 lead} and \emph{miss body} errors occur when a system skips a sentence
 containing a novel nugget in the lead or article body respectively. An
 \emph{empty} error indicates an update was selected that contained no nugget.
 \emph{Duplicate} errors occur when an update contains nuggets but none are
 novel. 
 
  Overall, errors of the miss type are most common and suggest future
 development effort should focus on summary content identification.  About a
 fifth to a third of all system error comes from missing content in the lead
 sentence alone.
 
  After misses, empty errors (false positives) are the next largest source of
 error. \textsc{Cos} was especially prone to empty errors (41\% of its total
 errors). \textsc{Ls} is also vulnerable to empties (19.9\%) but after
 applying the similarity filter and restricting to first sentences, these
 errors can be reduced dramatically (to 1\%).
  
  Surprisingly, duplicate errors are a minor issue in our evaluation. This is
 not to suggest we should ignore this component, however, as efforts to
 increase recall (reduce miss errors) are likely to require more robust
 redundancy detection. 

\section{Conclusion} \label{sec:conclusion}

  In this paper we presented a fully online streaming document summarization
 system capable of processing web-scale data efficiently. We also demonstrated
 the effectiveness of ``learning to search'' algorithms for this task.  As
 shown in our error analysis, improving the summary content selection
 especially in article body should be the focus of future work. We would like
 to explore deeper linguistic analysis (e.g. coreference and discourse
 structures) to identify places likely to contain content rather than
 processing whole documents. 

\section{Acknowledgements}

We would like to thank Hal Daum{\'e} III for answering our questions about 
learning to search. 
The research described here was supported in part
by the National Science Foundation (NSF) under
IIS-1422863. Any opinions, findings and conclusions
or recommendations expressed in this paper
are those of the authors and do not necessarily reflect
the views of the NSF.

%% The file named.bst is a bibliography style file for BibTeX 0.99c
\bibliographystyle{named}
\bibliography{ijcai16-2615}

\end{document}